\documentclass[runningheads]{llncs}
\usepackage[T1]{fontenc}
\usepackage{graphicx}
\usepackage{latexsym}
\usepackage{amssymb}
\usepackage{amsmath}
\usepackage{booktabs}
\usepackage{enumitem}
\usepackage{graphicx}
\usepackage{color}
\usepackage{algorithm}
\usepackage{algpseudocode}
\usepackage[square,sort,comma,numbers]{natbib}
\usepackage{subfigure} 
\usepackage{hyperref}
\usepackage{multirow} 
\begin{document}
\title{TempoKGAT: A Novel Graph Attention Network Approach for Temporal Graph Analysis}
\titlerunning{TempoKGAT}
%
\author{Lena Sasal\inst{1} \and
Daniel Busby\inst{2} \and
Abdenour Hadid\inst{1}}
\authorrunning{L. Sasal et al.}
%
\institute{Sorbonne Center for Artificial Intelligence, Sorbonne University, Abu Dhabi, UAE \email{lena.sasal@sorbonne.ae}  \and
TotalEnergies, France}
%
\maketitle   

\begin{abstract}
Graph neural networks (GNN) have shown significant capabilities in handling structured data, yet their application to dynamic, temporal data remains limited.
This paper presents a new type of graph attention network, called TempoKGAT, which combines time-decaying weight and a selective neighbor aggregation mechanism on the spatial domain, which helps uncover latent patterns in the graph data.
In this approach, a top-k neighbor selection based on the edge weights is introduced to represent the evolving features of the graph data.
We evaluated the performance of our TempoKGAT on multiple datasets from the traffic, energy, and health sectors involving spatio-temporal data. We compared the performance of our approach to several state-of-the-art methods found in the literature on several open-source datasets. Our method shows superior accuracy on all datasets.
These results indicate that TempoKGAT builds on existing methodologies to optimize prediction accuracy and provide new insights into model interpretation in temporal contexts. 
\end{abstract}


\section{Introduction}

Forecasting and predictive analysis have long been crucial in various domains, including economics, meteorology, and more recently, social and biological systems~\citep{surveyForecasting2016}. Traditionally, forecasting methods have relied on sophisticated statistical models \citep{hyndman2008ARIMA} primarily designed for time-series data within Euclidean spaces. However, the advent of deep learning has revolutionized traditional machine learning approaches by harnessing data with non-linear dependencies and complex interrelationships \citep{sasal2022Wtransformer}.

Graph Neural Networks (GNN) represent one of the most significant innovations in this new era, providing powerful tools for analyzing data structured as graphs \citep{scarselli2008graph}. The shift to GNN marked a transition from conventional deep learning applied to Euclidean spaces to a focus on non-Euclidean domains, such as social networks, biological ecosystems, and infrastructure networks \citep{zhou2020graph}.

\begin{figure}
    \centering
    \includegraphics[width=1.0\textwidth]{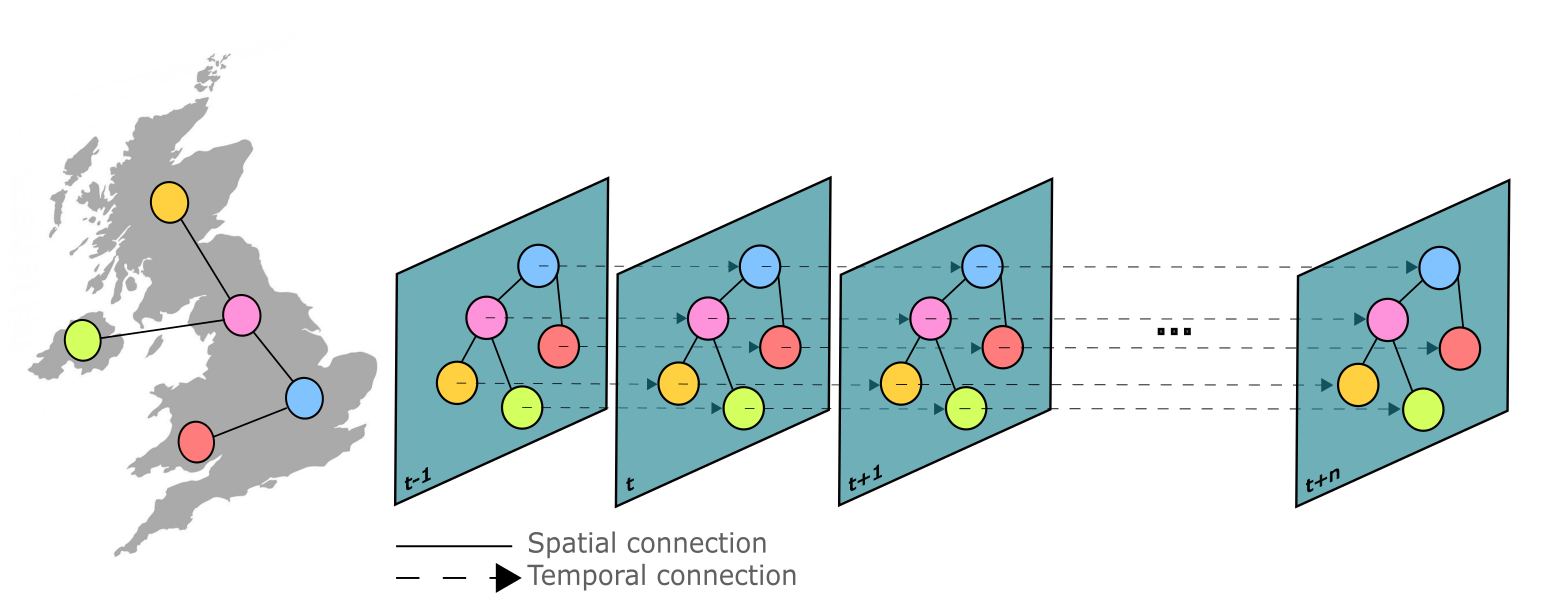}
    \caption{Example of a spatio-temporal graph's complexity, capturing evolving connections over time and space.}
    \label{fig:example}
\end{figure}

Despite the success of GNN, their application to temporal graphs, which are dynamic and involve constantly changing relationships as depicted in Fig.~\ref{fig:example}, poses significant challenges \citep{holme2012temporal}. These graphs are vital for applications like tracking disease spread \citep{salathe2010high}, urban planning \citep{zheng2014urban}, and traffic forecasting \citep{Jiang_2022Trafficsurvey}. There is a pressing need for models that not only understand static structures but also adapt to evolving relationships.

While Graph Attention Networks (GAT) have shown promise in addressing several challenges by dynamically weighting the importance of neighboring nodes~\citep{velickovic2018graph}, the unique characteristics of temporal graphs, including their dynamic network structures and edge properties, still demand the formulation of novel solutions. Although substantial progress has been made in developing GNN that can model temporal patterns \citep{gilmer2017neural, hamilton2017inductive}, the survey on dynamic graph neural networks \citep{zheng2024surveydynamicgraphneural} demonstrates that the field continues to evolve with significant ongoing research. This indicates an active area of development where further innovations can provide substantial contributions. This context motivates our work on TempoKGAT, which aims to leverage these advancements and address specific challenges in temporal graph analysis.

Moreover, the potential of edge attributes in GNN frameworks, particularly in time-varying graphs, remains underexploited \citep{gong2019exploiting}. Traditional GAT often overlooks the rich informational content of edge attributes by setting attention coefficients to zero between unconnected nodes. This oversight highlights the opportunity for significant improvements in GNNs' ability to model complex interactions more accurately, especially for graphs undergoing dynamic changes.

To address these challenges, this paper introduces TempoKGAT, a new methodology that extends GAT for effective temporal graph analysis. By incorporating time-decaying weights and optimizing neighborhood selection, TempoKGAT is designed to more accurately capture the evolving nature of graphs, thus elevating both predictive performance and model interpretability.

The key contributions of this work are outlined as follows:
\begin{itemize}
    \item We present TempoKGAT, an extension of the Graph Attention Network that integrates temporal dynamics through time-decay weighting and spatial information by selectively aggregating features from the most significant neighbors. This approach effectively captures the evolving nature of graph-structured data.
    \item We conducted extensive experiments on our model against a variety of state-of-the-art models on 4 different datasets and achieved better performance in all the cases based on 3 different metrics.
\end{itemize}

All the data and codes are available at \url{https://github.com/CapWidow/TempoKGATCode} for public use to support the principle of
reproducible research.

\section{Related Work}

The landscape of forecasting has evolved from classical statistical methods to advanced machine learning techniques, initially targeting time series data but increasingly applying graph-based approaches to leverage spatial dependencies~\citep{wu2016shortterm}. This shift to graph-based forecasting has seen adaptations of neural network architectures to accommodate graph structures, moving from Convolutional Neural Networks (CNN)~\citep{lecun1998cnn} to Graph Convolutional Networks (GCN)~\citep{kipf2017semisupervised} and further to Graph Attention Networks (GAT)~\citep{velickovic2018graph}.

Current models like Diffusion Convolutional Recurrent Neural Networks \\ (DCRNN) \citep{li2018dcrnn_traffic} and Temporal Graph Convolutional Networks (TGCN) \citep{zhao2019tgcnprediction} introduce innovative solutions incorporating temporal dynamics into graph structures. Yet, these models have not fully capitalized on the advantages that weighted edges can provide \citep{li2018adaptive, wang2019dynamic, seo2016structured, yu2017spatio}. Evolve GCN \citep{pareja2020evolvegcn} addresses the challenge of applying GCN to dynamic or evolving graphs by allowing the GCN parameters to evolve over time but still doesn't fully take advantage of edge features in the forecasting mechanism.

While substantial advancements in edge modeling have been documented in GNN, particularly in static and multi-relational contexts, the application of these advancements to temporal graph scenarios has not been as extensively explored. The models such as those described in \citep{schlichtkrull2017modelingrelationaldatagraph} and  \citep{vashishth2020compositionbasedmultirelationalgraphconvolutional} have significantly enhanced edge modeling by accommodating multiple types of relations and complex dependencies. However, when it comes to temporal graphs, the dynamic nature of edge information and its temporal relevance present unique challenges that are still being addressed. This underscores the motivation for our work with TempoKGAT, aiming to merge the benefits of advanced edge modeling techniques with the specific needs of temporal dynamics.

Our work seeks to build on these foundations by enhancing the role of edge weights in graph-based forecasting. We propose a refined application of edge attributes, combined with the temporal and spatial modeling capabilities of GNN, to significantly improve prediction accuracy. This approach not only addresses the limitations identified in prior research but also provides a sophisticated method for analyzing the complexities of temporal graph-structured data.
\section{Proposed TempoKGAT Model}

\begin{figure*}
    \centering
    \includegraphics[width=1\textwidth]{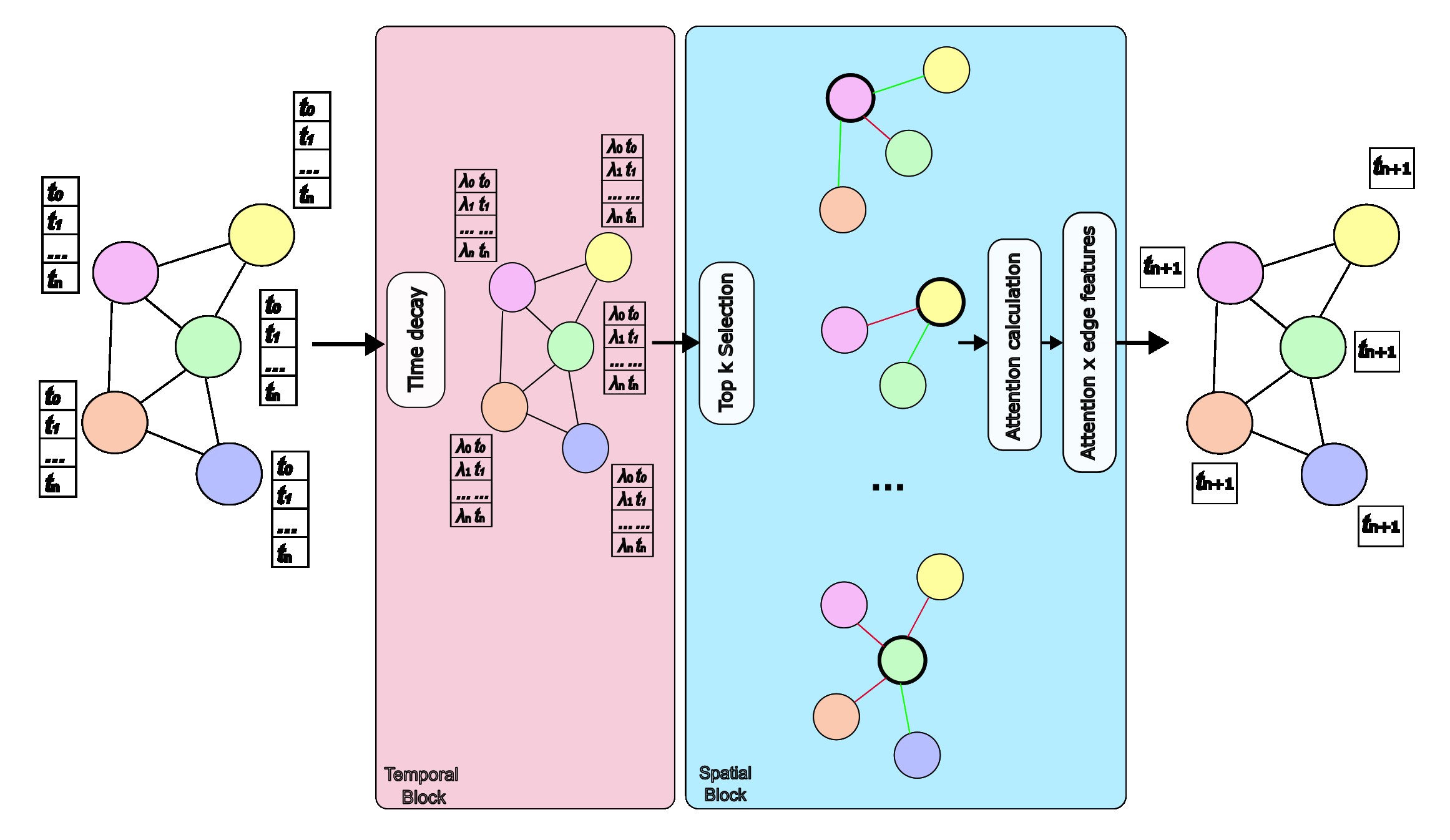}
    \caption{Diagrammatic Representation of the TempoKGAT Layer Workflow: This diagram traces the sequence of operations within the TempoKGAT layer, starting with the initial node features. It initiates with the Temporal Block, where a time decay function is applied to nodes to prioritize recent data. The process advances to the Spatial Block, depicting the selection of top-k neighbors based on edge weights, indicated by green lines, while excluded edges are in red. Following this, the focus shifts to computing attention coefficients, which gauge the significance of each chosen neighbor. The workflow concludes with an attention-weighted feature update, integrating temporal and spatial insights to refine node features for the next timestep (\(t_{n+1}\)).}

    \label{fig:architecture}
\end{figure*}

Given a graph $G = (V, E)$ with nodes $v \in V$ and edges $(v, u) \in E$ associated with edge weights $w_{vu}$, in this section, we delineate the architecture of the TempoKGAT layer, our proposed temporal graph attention network layer. This layer processes node features, $\mathbf{X} \in \mathbb{R}^{N \times F}$, where $N$ denotes the number of nodes, and $F$ represents the feature dimensionality per node. We aim to distill these inputs into refined node representations, $\mathbf{H} \in \mathbb{R}^{N \times F'}$, with $F'$ being the dimensionality of the output features. The design of TempoKGAT explained in Fig.~\ref{fig:architecture}, is predicated on capturing the nuanced interplay between the inherent graph structure and the temporal evolution of node features.

\subsection{Nodes Features Transformation with Temporal Decay}

The foundational step in the TempoKGAT layer involves imposing a temporal decay on the node features, a technique designed to foreground recent information crucial for understanding dynamic graphs. Given the input node features, $\mathbf{X} = \{\vec{x}_1, \vec{x}_2, \ldots, \vec{x}_N\}$, where each $\vec{x}_i \in \mathbb{R}^F$, we apply temporal decay to derive $\mathbf{X}_{\text{decay}}$:

\begin{equation}
    \mathbf{X}_{\text{decay}} = \mathbf{X} \odot e^{-\lambda \mathbf{t}},
\end{equation}

Here, $\lambda$ represents the decay rate, influencing how swiftly information's relevance fades over time, and $\mathbf{t}$ encapsulates the temporal proximity of each node feature, with $\mathbf{X}_{\text{decay}}$ signifying the temporally adjusted features. An exponential decay function is employed to model the diminishing influence of past interactions over time. This choice is grounded in the function's smoothness and its continuous derivative, which are conducive to stable gradient-based optimization. Additionally, the exponential decay naturally captures the gradual attenuation of influence, aligning with the temporal dynamics observed in real-world data. This ensures that recent events are weighted more heavily, yet does not disregard historical data, providing a nuanced temporal context to the graph representation.

\subsection{Top-k Neighbor Selection Based on Edge Weights}

Following temporal adjustment, we refine the graph's structure by identifying for each node $i$, its most significant neighbors through a top-k selection mechanism, prioritizing those with the strongest edge weights:

\begin{equation}
    \mathcal{N}_k(i) = \underset{j \in N(i)}{\mathrm{arg\,top\text{-}k}}\; (w_{ij}),
\end{equation}

where $\mathcal{N}_k(i)$ represents the chosen top-k neighbors of node $i$, ensuring that the attention mechanism's focus is directed towards structurally pertinent neighbors, thereby streamlining the computational emphasis towards the most meaningful node interactions

\subsection{Attention Mechanism and Weighted Aggregation}

Before delving into the specifics of our model, it's important to understand the fundamental concept of the attention mechanism that inspires our approach. The general form of an attention function can be expressed as:

\begin{equation}
\text{Attention}(Q, K, V) = \text{softmax}\left(\frac{QK^T}{\sqrt{d_k}}\right)V
\label{eq:general_attention}
\end{equation}

Here, $Q$, $K$, and $V$ represent the query, key, and value matrices respectively, and $d_k$ denotes the dimensionality of the keys. This formulation uses the softmax function to normalize the weights across the different queries, allowing the model to distribute its focus proportionally among the most informative parts of the input data. Such mechanisms are crucial in many modern neural architectures, facilitating selective attention over elements of the data that are most relevant to the task.

In the specific application within our TempoKGAT model, the attention mechanism has been specially adapted to handle graph-structured data. The process of dynamically weighting the influence of each node $i$'s neighbors begins by computing the raw attention coefficients, $e_{ij}$, for each neighbor $j$. This is achieved through a parameterized computation involving the node features, where the attention mechanism, represented by $\mathbf{a}$, operates on the transformed features of both the node and its neighbor:

\begin{equation}
    e_{ij} = \mathbf{a} \left( \mathbf{W}\vec{x}_{\text{decay}_i}, \mathbf{W}\vec{x}_{\text{decay}_j} \right),
\end{equation}

Here, $e_{ij}$ represents the preliminary attention score between node $i$ and its neighbor $j$, before any normalization. These scores are computed by applying a linear transformation $\mathbf{W}$ to the decay-adjusted features $\vec{x}_{\text{decay}}$, and then passing the concatenated result through the attention mechanism $\mathbf{a}$.

To ensure the attention scores are comparable across different nodes and to focus on the most relevant neighbors, we normalize these raw scores using the softmax function, leading to the final attention coefficients $\alpha_{ij}$. The softmax operation is applied across all neighbors of node $i$, which allows us to interpret $\alpha_{ij}$ as the probability that node $i$ attends to neighbor $j$:

\begin{equation}
    \alpha_{ij} = \frac{\exp\left( \text{LeakyReLU} \left( \vec{\mathbf{a}}^{\top} \left[ \mathbf{W}\vec{\mathbf{x}}_{\text{decay}_i} \| \mathbf{W}\vec{\mathbf{x}}_{\text{decay}_j} \right] \right) \right)}{\sum_{k \in \mathcal{N}_i} \exp\left( \text{LeakyReLU} \left( \vec{\mathbf{a}}^{\top} \left[ \mathbf{W}\vec{\mathbf{x}}_{\text{decay}_i} \| \mathbf{W}\vec{\mathbf{x}}_{\text{decay}_k} \right] \right) \right)},
\end{equation}

This formula applies the LeakyReLU function to introduce non-linearity and enhance the model's ability to learn complex patterns. The numerator calculates the exponentiated, transformed attention score for the neighbor $j$, while the denominator is a normalization term that sums over all exponentiated, transformed scores for all neighbors $k$ in $\mathcal{N}_i$, the neighborhood of node $i$. This ensures that $\alpha_{ij}$ sums to 1 over all neighbors $j$.

The normalized attention coefficients $\alpha_{ij}$ are crucial for the next step in our mechanism, the weighted aggregation of neighbor features. They allow the model to weigh each neighbor's features based on their relevance, as determined by the attention mechanism:

\begin{equation}
    \alpha_{ij} = \text{softmax}(e_{ij}),
\end{equation}

This equation succinctly expresses the normalization process described above, encapsulating how each attention score $e_{ij}$ is transformed into a coefficient $\alpha_{ij}$ that dictates the extent of influence neighbor $j$ has on node $i$.
Once we have calculated the normalized attention coefficients $\alpha_{ij}$, the next crucial step in our model is to aggregate the features of the neighbors to update the feature representation of each node. This aggregation takes into account both the attention score and the edge features of connecting our nodes $i$ to the neighbor node $j$:

\begin{equation}
    \beta_{ij} = \alpha_{ij} \cdot w_{ij},
\end{equation}

In this formula, $\beta_{ij}$ represents the new feature vector for node $i$ after aggregating the contributions from the connection to its neighbors.

Moreover, for a more nuanced incorporation of the graph's temporal dynamics, we also consider the weighted attention in conjunction with the edge weights. This results in the final aggregation formula:

\begin{equation}
    \vec{x'}_i = \sum_{j \in \mathcal{N}_i} (\beta_{ij} \cdot \vec{x}_{\text{decay}_j}),
\end{equation}

Here, $w_{ij}$ denotes the edge weights between node $i$ and its neighbor $j$, introducing an additional layer of relevance based on the strength of the connection between nodes. This approach ensures that the aggregation is sensitive not only to the attention mechanism's output but also to the inherent structure of the graph as encoded by the edge weights. This combined strategy allows TempoKGAT to dynamically adapt to both the structural and temporal aspects.

\begin{algorithm}
\caption{TempoKGAT Layer}
\begin{algorithmic}[1]
\State \textbf{Input:} Node features $X \in \mathbb{R}^{N \times F}$, edge indices $E$, edge weights $w$, parameter $k$, decay rate $\lambda_{decay}$
\State \textbf{Output:} Updated node features $X'$

\Procedure{TempoKGAT Layer}{$X, E, w$}
    \State $W \gets \text{Learnable weight matrix}$  
    \State $a \gets \text{Learnable attention coefficient vector}$  
    \State $LeakyReLU \gets \text{Leaky ReLU activation function}$  

    \State $time\_decays \gets \exp(-\lambda_{decay} \cdot \text{arange}(0, N))$  
    \State $X_{decay} \gets X \odot time\_decays$  

    \State $X' \gets \text{Zero matrix of size } X$  

    \For{each node $i$ in $1 \ldots N$}
        \State $(\text{neighbors}, \text{indices}) \gets \text{select\_top\_k\_neighbors}(i, E, w, k)$  
        \For{each $j$ in neighbors}
            \State $e_{ij} \gets LeakyReLU(a^\top \cdot (W X_{decay}[i] \| W X_{decay}[j]))$  
            \State $\alpha_{ij} \gets \text{softmax}(e_{ij})$  
            \State $\beta_{ij} \gets \alpha_{ij} \cdot w[\text{indices}[j]]$  
            \State $X'[i] \gets X'[i] + \beta_{ij} \cdot W X_{decay}[j]$  
        \EndFor
    \EndFor
    \State \Return $X'$
\EndProcedure

\Function{select\_top\_k\_neighbors}{$i, E, w, k$}
    \State Identify edges $(i, j)$ from $E$ and corresponding weights from $W$ for node $i$
    \State Sort the neighbors $j$ based on weights in descending order
    \State Select the top-$k$ neighbors and their indices in $W$
    \State \Return selected neighbors and their indices
\EndFunction
\end{algorithmic}
\end{algorithm}

\section{Experimental Setup}

In this section, we present a set of experiments to demonstrate the effectiveness of TempoKGAT. The
setting includes a variety of datasets, baselines, and evaluation metrics. For a fair comparison across all baselines, the hyperparameters are set to the same value.

\subsection{Dataset}

Our study leverages a diverse array of datasets, each uniquely suited to evaluating the performance of graph neural network models. These datasets differ in complexity, periodicity of data collection, and graph structure, providing a comprehensive test bed for our analyses. Detailed specifications, including node and edge counts, data periodicity, total observations, and node feature information, are summarized in Table~\ref{tab:datasets_description}.

\textbf{PedalMe}:
The PedalMe Bicycle delivery dataset, encompassing over 30 weekly snapshots from London (2020-2021), serves as a StaticGraphTemporalSignal in the PyTorch Geometric Temporal framework \citep{rozemberczki2021pytorch}. It features a static graph with localities as vertices (15 nodes), spatial connections as edges (225 edges), and vertex features including 4 lagged weekly delivery counts, aiming to predict the next week's delivery demand.

\textbf{England Covid}:
The England Covid dataset maps daily mobility and COVID-19 cases in England's NUTS3 regions from March 3 to May 12, through a directed, weighted graph based on Facebook's mobility data. Node features indicate regional COVID-19 case counts over a past window, aiming to predict next-day cases using Transfer Graph Neural Networks, as explored in the associated paper on pandemic forecasting. This dataset contains 129 nodes and 2158 edges and is dynamic throughout the timestep.

\textbf{Windmill}:
The dataset captures hourly energy output from windmills in a European country over a span exceeding two years, presented in two versions to elucidate the influence of graph size—small and medium—on analysis outcomes. The small graph version comprises 11 windmills and the medium contains 26, with vertices representing individual windmills and weighted edges indicating the strength of relationships between them. These variations enable regression tasks, providing a unique opportunity to study the impact of node and edge quantities on predictive modeling and network analysis efficacy.

\textbf{ChickenPox}:
The dataset comprises over 500 weekly snapshots of county-level chickenpox cases in Hungary, recorded from 2004 to 2014 and designed for use with PyTorch Geometric Temporal. It features a static graph where 20 counties are vertices connected by 102 edges representing their adjacency. Each vertex has features including 4 lagged weekly counts of cases, aiming to predict the number of cases for the next week.

\begin{table}[htbp]
\centering
\scriptsize
\caption{Description of Datasets Used in the Study}
\label{tab:datasets_description}
\begin{tabular}{@{}llllllll@{}}
\toprule
Dataset       & Nodes & Edges  & Observations & Node Features &Average Node Degree & Temporal Nature \\ \midrule
PedalMe       & 15    & 225         & 30  &  4 &15& Static\\
England Covid & 129   & 2158         & 52  &  8 &16& Dynamic\\
Small Windmill& 11    & 121         & 500 & 8 & 11 &Static \\
Medium Windmill&26 & 676       & 500 & 8 & 26 &Static \\
ChickenPox    & 20    & 102         & 516        & 4 & 5 &Static \\ \bottomrule
\end{tabular}
\end{table}

\subsection{Baseline Models}

We compare our model against several baseline models, each utilizing different mechanisms to integrate temporal and spatial dimensions into graph-structured data analysis:

\textbf{GRU (Graph GRU)}: Graph GRU extend the capabilities of traditional Gated Recurrent Units by integrating them into graph structures. This is achieved by applying GRU operations on the node embeddings, which are updated based on the neighborhood features aggregated during each step. This model is particularly effective in capturing temporal dynamics along with maintaining the graph topology, which is vital for tasks like dynamic network analysis.

\textbf{LSTM (Graph LSTM)}: Graph LSTM modify Long Short-Term Memory networks to handle graph data by incorporating LSTM units in the graph propagation process. Node features in a graph are updated through LSTM units, where the inputs are the aggregated features of the neighbors, allowing the model to capture both long-term and short-term temporal patterns in graph-structured data.

\textbf{GCN (Graph Convolutional Networks)}: Graph Convolutional Networks use a specialized form of convolution adapted for graphs. This involves the multiplication of a graph’s adjacency matrix with the node features matrix followed by a non-linear transformation. This process aggregates neighboring node features into a compact representation, which helps in capturing local graph topology effectively.

\textbf{GAT (Graph Attention Networks)}: Graph Attention Networks employ a self-attention strategy to weigh the importance of each node's neighbors. By computing attention coefficients that signify the importance of each edge during the feature aggregation phase, GAT can adaptively focus on the most relevant parts of the data, enhancing model performance on complex graph structures.

\textbf{TGCN (Temporal Graph Convolutional Networks)}: TGCN add a temporal dimension to GCN by incorporating time-stepped graph convolutions. This model handles dynamic graphs where the connectivity and edge weights can change over time, using convolutional layers that are applied separately at each time step, allowing it to recognize and forecast evolving patterns.

\textbf{DCRNN (Diffusion Convolutional Recurrent Neural Networks)}: DCRNN merge graph convolution with recurrent neural architecture to capture spatial-temporal patterns. The model uses a diffusion process as a proxy for the graph convolution operation, which mimics the spread of information across the graph. Coupled with recurrent units, it effectively captures dynamic changes over both space and time.

\textbf{EvolveGCNH (Evolve Graph Convolutional Neural Networks with Highway)}: EvolveGCN extend traditional GCN by allowing the convolutional filters to evolve, which adapts to the dynamic nature of the graphs. This is coupled with highway networks that provide paths for training signals to travel at different depths, enhancing training efficiency and allowing the network to learn complex temporal patterns in the graph data.

\subsection{Evaluation Metrics}

To quantify the performance of our models, we employ the following metrics:

\textbf{Mean Squared Error (MSE)} measures the average of the squares of the errors. It is calculated as:
\begin{equation}
    MSE = \frac{1}{n} \sum_{i=1}^{n} (Y_i - \hat{Y}_i)^2
\end{equation}

\textbf{Root Mean Squared Error (RMSE)} is the square root of MSE, which adjusts the scale of the errors to be compatible with the original data points:
\begin{equation}
    RMSE = \sqrt{MSE} = \sqrt{\frac{1}{n} \sum_{i=1}^{n} (Y_i - \hat{Y}_i)^2}
\end{equation}

\textbf{Mean Absolute Error (MAE)} calculates the average of the absolute differences between the predicted values and observed values, offering a direct measure of average error:
\begin{equation}
    MAE = \frac{1}{n} \sum_{i=1}^{n} |Y_i - \hat{Y}_i|
\end{equation}

Where $n$ is the number of observations, $Y_i$ is the actual value, and $\hat{Y}_i$ is the predicted value for the $i$-th observation.\\\\

All models in this study were subjected to a standardized experimental setup to ensure comparability across different tests. The core architecture of each model involves a sequence of operations beginning with feature transformation via the chosen model, followed by ReLU activation and a linear transformation. This configuration allows us to evaluate the fundamental aspects of each model's ability to process and learn from graph-structured data.

The optimization is done using the Adam optimizer with a learning rate of \(0.001\). This optimizer is well-suited for our experiments due to its efficient handling of sparse gradients and adaptive learning rate capabilities. The Mean Squared Error (MSE) was employed as the loss function. MSE is particularly useful in regression tasks, which aligns with the objectives of our datasets where the goal is to predict numerical values accurately.

Each model was trained over 200 epochs to balance between adequate learning and avoiding overfitting. The dataset was divided into training and testing sets using an 80/20 split. This division was facilitated by the \\ \texttt{temporal\_signal\_split} function from PyTorch Geometric, ensuring that each model was trained and validated on distinct sets of data, thus providing a robust assessment of their generalization capabilities.

Our experimental design guarantees the fairness of the performance comparison among the different models and also contributes to the reproducibility of the results, which consequently facilitates further research and validation.

\section{Results and Discussion}

\renewcommand{\arraystretch}{1.3}
\begin{table*}[!t]
\centering
\caption{Model Performance Across Multiple Datasets with and without Edge Feature. For TempoKGAT, the $k$ value, used in the top $k$ selection, is for the lowest evaluation metrics value for each dataset (k = 1 for PedalMe, k = 1 for Chicken Pox, k = 5 for England Covid, k = 7 for Small WindMill and k = 17 for Medium WindMill) .}
\label{table:merged_model_performance}
\resizebox{\textwidth}{!}{%
\begin{tabular}{|l|ccc|ccc|ccc|ccc|ccc|}
\hline
\multirow{2}{*}{\textbf{Model}} & \multicolumn{3}{c|}{\textbf{PedalMe}} & \multicolumn{3}{c|}{\textbf{ChickenPox}} & \multicolumn{3}{c|}{\textbf{England Covid}} & \multicolumn{3}{c|}{\textbf{Small WindMill}} & \multicolumn{3}{c|}{\textbf{Medium WindMill}} \\
 & \textbf{MAE} & \textbf{MSE} & \textbf{RMSE} & \textbf{MAE} & \textbf{MSE} & \textbf{RMSE} & \textbf{MAE} & \textbf{MSE} & \textbf{RMSE} & \textbf{MAE} & \textbf{MSE} & \textbf{RMSE} & \textbf{MAE} & \textbf{MSE} & \textbf{RMSE} \\
\hline
GRU  \citep{seo2016structured}         & 0.8337 & 1.3445 & 1.1595 & 0.6989 & 1.1104 & 1.0538 & 0.7031 & 0.7125 & 0.8441 & 1.2901 & 2.5399 & 1.5937 & 1.0887 & 1.9138 & 1.3834 \\
GRU w edge\_weights                    & 0.8872 & 1.4899 & 1.2206 & 0.6989 & 1.1112 & 1.0541 & 0.6981 & 0.6977 & 0.8353 & 1.1143 & 1.8709 & 1.3678 & 0.9545 & 1.5025 & 1.2258 \\
LSTM  \citep{chen2021gclstm}           & 0.8412 & 1.2508 & 1.1184 & \textit{0.6511} & \textit{1.0062} & \textit{1.0031} & 0.6532 & 0.6044 & 0.7774 & 1.3174 & 2.6557 & 1.6296 & 1.0785 & 1.7948 & 1.3397 \\
LSTM w edge\_weights                   & 0.7884 & \textit{1.2131} & \textit{1.1014} & 0.6515 & 1.0123 & 1.0061 & \textit{0.6485} & 0.5925 & 0.7697 & 1.0225 & 1.6257 & 1.2750 & 1.0030 & 1.6266 & 1.2754 \\
EvolveGCNH  \citep{pareja2020evolvegcn}& 0.9656 & 1.4952 & 1.2228 & 0.6591 & 1.0551 & 1.0272 & 0.8941 & 1.0781 & 1.0383 & 1.2806 & 2.8041 & 1.6745 & 1.0633 & 2.2262 & 1.4921 \\
EvolveGCNH w edge\_weights             & 0.8457 & 1.3510 & 1.1623 & 0.6657 & 1.0599 & 1.0295 & 0.7708 & 0.8053 & 0.8974 & 1.1834 & 2.3063 & 1.5186 & 1.0483 & 2.0292 & 1.4245 \\
GCN   \citep{kipf2017semisupervised}   & 0.8902 & 1.4469 & 1.2029 & 0.6928 & 1.1017 & 1.0496 & 0.6514 & \textit{0.5791} & \textit{0.7610} & 0.9515 & \textit{1.2661} & \textit{1.1252} & \textit{0.8137} & \textit{1.0643} & \textit{1.0317} \\
GCN w edge\_weights                    & 0.8276 & 1.2703 & 1.1271 & 0.6912 & 1.0991 & 1.0484 & 0.7753 & 0.7646 & 0.8744 & \textit{0.9252} & 1.3134 & 1.1460 & 0.8406 & 1.1507 & 1.0727 \\
DCRNN  \citep{li2018dcrnn_traffic}     & 0.7835 & 1.2648 & 1.1246 & 0.6719 & 1.0721 & 1.0354 & 0.8445 & 0.9463 & 0.9728 & 0.9585 & 1.4180 & 1.1908 & 0.8403 & 1.2044 & 1.0974 \\
DCRNN w edge\_weights                  & \textit{0.7791} & 1.2489 & 1.1176 & 0.6749 & 1.0756 & 1.0371 & 0.8575 & 0.9716 & 0.9857 & 1.0035 & 1.6112 & 1.2693 & 0.8376 & 1.1867 & 1.0894 \\
TGCN \citep{zhao2019tgcnprediction}    & 0.8079 & 1.2807 & 1.1317 & 0.7070 & 1.1298 & 1.0629 & 0.7078 & 0.7075 & 0.8411 & 1.2437 & 2.4135 & 1.5535 & 1.0769 & 1.7561 & 1.3252 \\
TGCN w edge\_weights                   & 0.8302 & 1.3568 & 1.1648 & 0.7031 & 1.1213 & 1.0589 & 0.8463 & 0.9505 & 0.9750 & 1.1145 & 1.9032 & 1.3796 & 0.9922 & 1.6791 & 1.2958 \\
GAT \citep{velickovic2018graph}        & 1.0085 & 1.5636 & 1.2504 & 0.6933 & 1.1008 & 1.0492 & 0.7228 & 0.7238 & 0.8508 & 0.9589 & 1.3958 & 1.1814 & 0.8242 & 1.1194 & 1.0580 \\
GAT w edge\_weights                    & 1.0948 & 1.8247 & 1.3508 & 0.6894 & 1.0945 & 1.0462 & 0.6981 & 0.6869 & 0.8288 & 1.0242 & 1.7006 & 1.3041 & 0.8821 & 1.2601 & 1.1225 \\
\textbf{TempoKGAT} & \textbf{0.7476} & \textbf{1.1717} & \textbf{1.0825} & \textbf{0.6489} & \textbf{1.0017} & \textbf{1.0008} & \textbf{0.4953} & \textbf{0.4192} & \textbf{0.6474} & \textbf{0.7949} & \textbf{0.9821} & \textbf{0.9910} & \textbf{0.7198} & \textbf{0.8890} & \textbf{0.9429} \\
\hline
\end{tabular}%
}
\end{table*}

\begin{figure}
    \centering
    \includegraphics[width=1.0\textwidth]{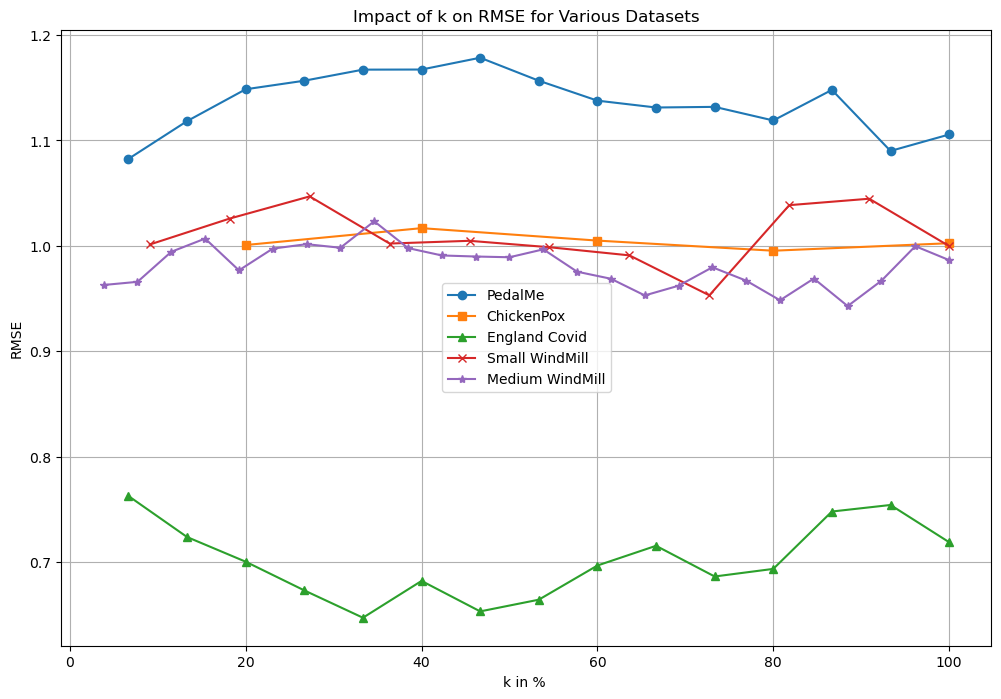}
    \caption{Impact of the \( k \) parameter in the top \(k\) selection on the RMSE across different datasets. Each curve represents a unique dataset, showcasing how variations in \( k \) in percentage, where for each dataset the minimum \( k \) is 1 and the maximum \( k \) is the average node degree of the dataset, affect the predictive accuracy of the model. This analysis illustrates the significance of choosing an optimal \( k \) value for enhancing TempoKGAT performance across diverse graph-based tasks.}
    \label{fig:rmse_k_impact}
\end{figure}

TempoKGAT's distinct capability to adapt its temporal and spatial understanding of data is demonstrated by its exceptional performance across various datasets. Table  ~\ref{table:merged_model_performance} showcase the predictive prowess of TempoKGAT, highlighting its significant benefits from strategically tuning its neighborhood size parameter, \(k\), as displayed in Fig.~\ref{fig:rmse_k_impact}. This model utilizes unique components that effectively integrate temporal changes and network dynamics, crucial for maximizing the potential of graph-based data.

The performance enhancements seen in the PedalMe and ChickenPox datasets are a testament to the novel capabilities of TempoKGAT. In the complex network of PedalMe, employing an optimal \(k=1\) with TempoKGAT not only builds on existing methods but also significantly shifts the performance metrics, with increases in MAE by 31.71\%, MSE by 35.78\%, and RMSE by 19.87\% compared to original GAT. This illustrates TempoKGAT's unique ability to capitalize on critical features from immediate neighbors and emphasizes the importance of recent interactions within complex network structures. Similarly in the ChickenPox dataset, even with its simpler graph structure, the adoption of TempoKGAT leads to notable performance shifts, improving original GAT by 5.87\% for MAE, 8.49\% for MSE, and 4.34\% for RMSE. This reflects how the model’s edge weighting innovatively enhances the impact of node interactions, maintaining efficacy across varying levels of network complexity.

In confronting the dynamic and evolving patterns of the England Covid dataset, TempoKGAT adapts effectively at \(k=5\) with improvements of 29.04\% in MAE, 38.97\% in MSE, and 21.88\% in RMSE of the original GAT implementation. The adaptability shown here illustrates the model's capacity to fully harness temporal dynamics and adjust to frequently changing data.

For the WindMill datasets, the strategic selection of optimal \(k\) values is crucial. In the Small WindMill dataset, TempoKGAT demonstrates its innovative approach by achieving a 22.38\% improvement in MAE, 42.25\% in MSE, and 24.01\% in RMSE against GAT. In the Medium WindMill dataset, the model introduces notable advancements, compared to original GAT, with an 18.39\% improvement in MAE, 29.45\% in MSE, and 16.01\% in RMSE. The performance peaks at approximately 64\% and 65\% of the respective maximum \(k\) values for each dataset, underscoring TempoKGAT's scalable and adaptable design in handling different dataset sizes.

A consistent observation across all datasets is the exemplary performance of TempoKGAT when set at \(k=1\), where it outstrips all other models regardless of edge weights integration. This indicates the model’s inherent strength in capturing essential features from immediate neighbors alone, offering a strong foundation for further enhancements through fine-tuning of the \(k\) value.

The  superior performance of TempoKGAT in forecasting, as depicted across these datasets, validates its sophisticated capabilities in effectively utilizing both the temporal and spatial aspects of graph data. Its robustness and versatility shine across diverse scenarios, from the intricate networks of PedalMe to the more straightforward structures found in ChickenPox, and in the temporally dynamic England Covid scenario. These results firmly establish TempoKGAT as a formidable tool for various predictive analytics applications within graph-based systems and lay the groundwork for continued advancements in the domain.

\section{Conclusion and Future Directions}

This paper introduced TempoKGAT, a novel graph attention network tailored for temporal graph analysis. By integrating time-decaying weights and selective neighbor aggregation, TempoKGAT has demonstrated superior predictive accuracy across various datasets, including PedalMe, ChickenPox, England Covid, and WindMill configurations. Notably, TempoKGAT achieves an average improvement in RMSE, MAE, and MSE metrics, outperforming traditional models in dynamic environments.

Despite its advantages, the computational demands of TempoKGAT increase with the neighborhood size parameter \(k\). Future work will focus on enhancing computational efficiency through algorithmic optimizations and exploring multi-head attention integration to capture more complex relational dynamics. Furthermore, the potential to scale TempoKGAT for larger graphs promises broader applications in domains requiring robust, responsive analytical tools. Ultimately, the development of TempoKGAT opens new avenues for innovation in graph-based predictive analytics.



\begin{credits}
\subsubsection{\ackname}
 The support of TotalEnergies is fully acknowledged. Lena Sasal (PhD Student) and Abdenour Hadid  (Professor,  Industry Chair at SCAI Center of Abu Dhabi) are funded by TotalEnergies collaboration agreement with Sorbonne University Abu Dhabi.


\bibliographystyle{splncs04}
\bibliography{3131}
\end{credits}
\end{document}